# State and Topology Estimation for Unobservable Distribution Systems using Deep Neural Networks

Behrouz Azimian, *Student Member, IEEE*, Reetam Sen Biswas, *Member, IEEE*, Shiva Moshtagh, *Student Member, IEEE*, Anamitra Pal, *Senior Member, IEEE*, Lang Tong, *Fellow, IEEE*, and Gautam Dasarathy, *Senior Member, IEEE*

*Abstract*— Time-synchronized state estimation for *reconfigurable* distribution networks is challenging because of limited real-time observability. This paper addresses this challenge by formulating a deep learning (DL)-based approach for topology identification (TI) *and* unbalanced three-phase distribution system state estimation (DSSE). Two deep neural networks (DNNs) are trained for *time-synchronized* DNN-based TI and DSSE, respectively, for systems that are incompletely observed by synchrophasor measurement devices (SMDs) in real-time. A data-driven approach for judicious SMD placement to facilitate reliable TI and DSSE is also provided. Robustness of the proposed methodology is demonstrated by considering non-Gaussian noise in the SMD measurements. A comparison of the DNN-based DSSE with more conventional approaches indicates that the DL-based approach gives better accuracy with smaller number of SMDs.

## I. INTRODUCTION

Real-time monitoring and control of distribution networks was traditionally deemed unnecessary because it had radial configuration, unidirectional power flows, and predictable load patterns. However, the fast growth of behind-the-meter (BTM) generation, particularly solar photovoltaic (PV), electric vehicles, and storage, is transitioning the distribution system from a passive load-serving entity to an active market-ready entity, whose reliable and secure operation necessitates real-time situational awareness [1]-[2]. Phasor measurement units (PMUs), distribution-PMUs (D-PMUs), and/or micro-PMUs, collectively referred to as synchrophasor measurement devices (SMDs) in this paper, have been introduced into the distribution system to provide fast (sub-second) situational awareness by enabling time-synchronized estimation [3]-[5]. However, the number of SMDs in a typical distribution network are not large enough to provide an *independent* assessment of the system state. The assumption of Gaussian noise in synchrophasor measurements has also been disproved recently [6]-[7].

At the same time, modern distribution systems are being equipped with advanced metering infrastructure (AMI) in the form of smart meters. By 2020, 100 million+ smart meters had been installed in the U.S. alone [8]. Hence, prior research has combined smart meter data with SMD data for facilitating distribution system state estimation (DSSE) [9]. However, smart meters measure energy consumption from 15 minute to hourly time intervals and report their readings after a few hours or even a few days [10]. These two aspects make smart meter data unsuitable for real-time DSSE. Moreover, smart meter data are not time-synchronized, which makes their direct integration with SMD data a statistical challenge.

To overcome the need for large numbers of real-time sensors, prior approaches for performing DSSE have used load forecasts as pseudo-measurements [11], [12]. However, it has been shown in [13] that using forecasted/pseudo-measurements in real-time can deteriorate estimation performance. Instead, [13] proposed a Bayesian approach that trained a deep neural network (DNN) to circumvent the real-time unobservability problem. However, the approach was not validated for three-phase unbalanced distribution systems. In [14], an artificial neural network was created for three-phase unbalanced DSSE. However, smart meter measurements were not considered in the analysis (only micro-PMU measurements were used) and loads were varied by a Gaussian distribution which might not correctly represent system behavior. A sparse-tracking state estimator for unbalanced distribution systems that are incompletely observed by D-PMUs was developed in [15]. However, it required additional information from event data that may not be always available and was restricted to radial networks. A three-phase DSSE based on a Bayesian fusion procedure was proposed in [16] to account for the different temporal aspects of the states and measurements. However, due to the heavy computational burden of the procedure it could not handle non-Gaussian loads and measurement noise. Moreover, in [11]-[16], the system topology was assumed to be fixed.

As topology of a distribution network changes with time, it is important to consider its impacts on DSSE [17]. In [18], mixed integer linear programming was used to estimate topology of distribution networks. However, the methodology required real-time measurements from line flow meters and smart meters, which are not available in most distribution systems. In [19], a graph-based optimization framework was proposed to recover topology of radial distribution networks using limited number of real-time meters. However, meshed grids and unbalanced multiphase distribution systems were not considered. In [20], a data-driven probabilistic network model was used for topology recognition. However, the method relied on smart meter data which made it unsuitable for real-time knowledge of network topology. In [21], a time-series signature verification method was used to track topology changes from streaming micro-PMU measurements. One switching at a time and prior information of the switch status were two assumptions that limited the usefulness of this method. In [22], a machine

This work was supported in part by the Department of Energy under grants DE-AR00001858-1631 and DE-EE0009355, by the Power Systems Engineering Research Center (PSERC) Grant T-63, and by the National Science Foundation (NSF) under the award OAC-1934766 and ECCS-2145063.
B. Azimian, R.S. Biswas S. Moshtagh, A. Pal, and G. Dasarathy are with the School of Electrical, Computer and Energy Engineering, Arizona State University, Tempe, AZ, 85287, USA (e-mail: bazimian@asu.edu; rsenbisw@asu.edu; smoshta1@asu.edu; apal12@asu.edu; gautamd@asu.edu)
L. Tong is with the School of Electrical and Computer Engineering, Cornell University, Ithaca, NY 14850, USA (e-mail: lt35@cornell.edu)

learning (ML)-based framework was proposed for topology identification (TI). However, the need for nodal currents, voltages, and power-factor angle of each phase limited its real-time applicability. In [23], a two-step numerical method was proposed to perform topology estimation. However, the method was too slow for real-time monitoring and was limited to balanced networks. Ref. [24] performed real-time state and topology estimation in unbalanced distribution networks. However, it used forecasted load data as pseudo-measurements, which can deteriorate its performance. Lastly, a systematic approach for identifying measurement locations that boosted estimation performance was not considered in [18]-[24].

This paper addresses the knowledge gaps identified above by making the following salient contributions:

1. A DNN-based TI is proposed to estimate switch statuses in real-time from sparsely placed SMDs.
2. A DNN-based DSSE for unbalanced three-phase distribution networks is developed that estimates states (voltage phasors) in a fast, time-synchronized manner for both radial and meshed networks.
3. Transfer learning is employed to account for the effects of topology changes on DNN-based DSSE.
4. A judicious approach for SMD placement to facilitate reliable TI and DSSE is presented.
5. Robustness of the proposed method is demonstrated by considering non-Gaussian noise in SMD measurements.

## II. MOTIVATION & THEORETICAL BACKGROUND

### A. Need for ML for Time-Synchronized DSSE

Time-synchronized state estimation in distribution networks using classical approaches, such as least-squares, requires the system to be completely observed by SMDs. However, it is highly unlikely that, at least in the near future, a distribution system will be equipped with as many SMDs as is required for complete real-time observability. To circumvent the problem of scarcity of SMDs for doing time-synchronized DSSE, a Bayesian approach is formulated in this paper in which the state, $x$, and the measurement, $z$, are treated as random variables. A minimum mean squared error (MMSE) estimator is then created to minimize the estimation error as shown below.

$$\min_{\hat{x}(z)} \mathbb{E}(\| \ x - \hat{x}(z) \ \|^2) \Longrightarrow \hat{x}^*(z) = \mathbb{E}(x|z) \qquad (1)$$

The MMSE estimator directly minimizes the estimation error while classical estimators, such as least-squares, minimize the modeling error embedded via a measurement function that relates the measurements with the states. By circumventing the need for the measurement function, the real-time observability requirements get bypassed in a Bayesian state estimator [13]. However, in (1), there are two underlying challenges to computing the conditional mean. First, the conditional expectation, which is defined by,

$$\mathbb{E}(x|z) = \int_{-\infty}^{+\infty} x p(x|z) dx \qquad (2)$$

requires the knowledge of $p(x, z)$, the joint probability density function (PDF) between $x$ and $z$. When the number of SMDs is scarce, the PDF between SMD data and all voltage phasors is unknown or impossible to specify, making direct computation of $\hat{x}^*(z)$ intractable. Second, even if the underlying joint PDF is known, finding a closed-form solution for (2) can be difficult.

A DNN is used here to approximate the MMSE state estimator as a DNN has excellent approximation capabilities [25]; i.e., the DNN for DSSE finds a mapping, $\mathcal{K}(\cdot)$, that relates $x$ and $z$.

### B. Transfer Learning

Now, a DNN can successfully approximate $\mathbb{E}(x|z)$ for a given topology. However, once the topology changes, the distribution of the inputs (i.e., SMD measurements) for which the DNN had been trained for, change. This is best realized from the fact that the direction of the currents in a feeder can reverse when topology change occurs. Thus, there is a need to update the trained DNN once the topology changes. One way to do this is to train the DNN for DSSE afresh for every new topology. However, doing so may take a long time. An alternate (better) solution is to use *Transfer learning* to transfer the knowledge gained from the old topology to the new topology.

Transfer learning tries to improve the learning of the target prediction function in the target domain using the knowledge available in the source domain and task. A domain $\mathcal{D}$ comprises two parts: a feature space, $\mathcal{Z}$, and a marginal probability distribution, $P(z)$. Given $\mathcal{D}$, a task $\mathcal{T}$ comprises two parts: a label space, $\mathcal{X}$, and an objective prediction (mapping) function, $\mathcal{K}(\cdot)$. In DNN-based DSSE under varying topologies, $\mathcal{Z}$ does not change as the same SMD measurements will be used for different topologies. However, $P(z)$ changes because loads are served by different paths when topology changes. i.e., $\mathcal{D}_S \neq \mathcal{D}_T$. Similarly, $\mathcal{X}$ does not change because the number of states (i.e., voltage phasor at each node) and their nature are the same. However, $\mathcal{K}(\cdot)$, must be retrained for the target domain, i.e., $\mathcal{T}_S \neq \mathcal{T}_T$. In accordance with this problem set-up, *inductive Transfer learning* [26] is applied in this paper to induce transfer of knowledge gained from $\mathcal{D}_S$ and $\mathcal{T}_S$ (old topology) to $\mathcal{D}_T$ and $\mathcal{T}_T$ (new topology).

Four approaches have been proposed for implementing inductive Transfer learning: feature-representation transfer, instance transfer, relational-knowledge transfer, and parameter transfer [27]. Here we use *parameter transfer* to update the DNN for DSSE as the DNN's parameters can be used for multiple domains. Two well-known parameter-based transfer learning methods are parameter-sharing and fine-tuning. Parameter-sharing assumes that the parameters are highly transferable due to which the parameters in the source domain can be directly copied to the target domain, where they are kept "frozen". Fine-tuning assumes that the parameters in the source domain are useful, but they must be trained with limited target domain data to better adapt to the target domain [28]. Since there is no guarantee that the parameters of the DNN-based DSSE will be highly transferable for different topologies, *fine-tuning* is used in this paper to update the weights of the DNN for DSSE when topology changes. Essentially, fine-tuning provides a more effective initialization (than random initialization) by using the weights from the previously well-trained DNN. By doing this, it bypasses the need for large amounts of data (and time) for DNN re-training (see Section V.D for implementation of the proposed methodology).

### C. Hyperparameter Tuning

One of the main challenges in DNN training is hyperparameter tuning. The hyperparameters that typically need tuning are batch size, dropout rate, number of hidden layers, number of neurons in each hidden layer, learning rate, and the optimizer. Hyperparameter tuning is usually done by grid search, Bayesian search, or random search. Grid search iterates over every combination of hyperparameter values and is therefore computationally very expensive. The Bayesian approach creates a probabilistic model of metric score as a function of the hyperparameters and chooses parameters with high probability of improving the metric. Although, it shows good performance for small number of continuous parameters, it does not scale to large number of different hyper-parameters. The Random search goes through different combinations of predefined sets of hyperparameters to identify the combination that gives the best result (lowest validation loss) [29]. As it has reasonable computational burden and good scalability, Random search was used in this paper for hyperparameter tuning.

## III. DNN Architecture for DSSE and TI

### A. DNN Architecture for DSSE

The basic structure of the proposed DNN is shown in Figure 1. Its inputs are the $z$ obtained from SMDs, the outputs are the estimated voltage phasors, $\hat{x}^*(z)$, $m$ refers to the size of $z$, $n$ refers to the total number of states to be estimated, $a$ denotes the activation function, $b$ denotes the bias, and $W$ refers to the weights conveying the output of previous neurons to the neurons of the next layer. Dropout is also applied to avoid overfitting; its effect is shown in Figure 1 by dotting some of the circles in the hidden layers. Note that for a distribution network that is incompletely observed in real-time, $n \gg m$. The number of neurons and hidden layers are hyperparameters that must be tuned offline. The rectified linear unit (ReLU) activation function is used for the hidden layers, while a linear activation function is used for the output layer. The loss function is chosen to be the empirical mean-square error which is consistent with the Bayesian approach. During the offline training process the weights are optimized to minimize the mean squared error using the backpropagation algorithm [30]. In real-time operation, SMD data is fed into the trained feed forward DNN and the estimated state, $\hat{x}^*$, is obtained.

### B. DNN Architecture for Topology Identification (TI)

The DNN for DSSE shown in Figure 1 is trained based on the assumption that the topology of the system is known and fixed. However, when a topology change occurs, this DNN, which is trained for the old topology, will receive test data from another feature space that corresponds to the new topology. As this might lower the performance of this DNN, a sequential procedure is adopted in which the new topology is identified first by a different DNN, and the DNN for DSSE is updated afterwards based on the identified (new) topology.

As opposed to the regression DNN that was built for DSSE, a classification DNN is built for DNN-based TI in which the measurements from sparsely placed SMDs are used to track the switch statuses in real-time. In the DNN for TI, the number of neurons in the output layer is equal to the number of feasible topologies in the network[1], the SoftMax function is used as the activation function for the output layer, while the categorical cross-entropy is chosen to be the loss function (the inputs and activation function for the hidden layers are the same as the DNN for DSSE). For training the DNN for TI, the database generation process (see Section V.A) is repeated for all feasible topologies. A distinct advantage of the proposed DNN-based TI is that it only requires high-speed time-synchronized SMD measurements for online operation as opposed to [18], [20], which needed smart meter measurements in real-time.

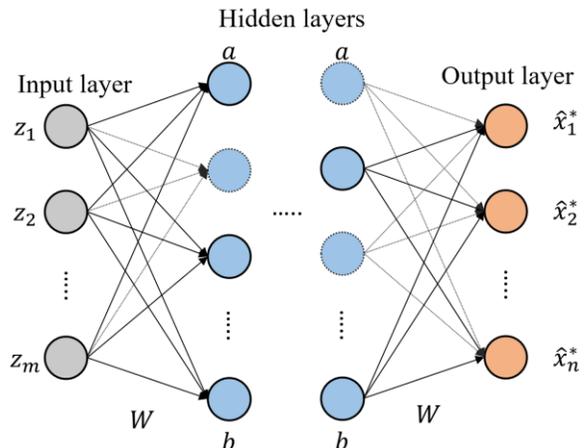

Figure 1: Basic DNN architecture for DNN-based DSSE with dropout

## IV. Measurement Selection

The proposed approach for DSSE and TI uses SMD data in real-time for state and topology estimation, respectively. Now, due to economical constraints, it is not viable to place many SMDs in the distribution system. Therefore, a systematic algorithm is needed to select locations for placing a small number of SMDs to obtain reliable and accurate results for both DSSE and TI. An integrated framework is proposed here to identify suitable locations for placing SMDs for performing DNN-based DSSE and TI. Since it is crucial to know the network model before doing DSSE, TI must be performed first to estimate the current network topology. Hence, we initially find the locations for accurate TI (Section IV.A). If those locations do not satisfy the criteria for measurement selection for DNN-based DSSE, we find additional locations where SMDs can be placed (Section IV.B).

### A. Measurement selection for DNN-based TI

DNN-based TI is a classification problem in which we estimate the topology of the system, i.e., the status of all the switches, from SMD measurements. Hence, measurement selection for DNN-based TI can be viewed as a *feature selection problem*, whose objective is to find the suitable location of SMDs required to achieve acceptable TI performance. Current phasors (in contrast to voltage phasors) are used for training the DNN classifier as opening/closing the switches will have a bigger influence on the currents flowing through the network. *Sequential forward selection* [31], a greedy search algorithm that starts with an empty set and adds features based on the ML classifier accuracy, is used to determine the appropriate current phasor measurements. The number of desired features is a hyperparameter that is tuned to find the required number of SMDs for a given accuracy level/budget constraint. For example, if α% accuracy is desired, then the number of desired features (SMDs) is increased gradually until an accuracy of α% is reached. However, if the budget constraint is violated first, then the number of SMDs placed before the budget was exceeded, is used to perform both DNN-based TI and DSSE.

---

[1] Feasible topologies refer to those switch configurations for which the system does not split into islanded sub-systems.

### B. Measurement selection for DNN-based DSSE

DNN-based DSSE is a regression problem for which all voltages and currents of the distribution network can be potential input features. The most common technique for finding the best features for a regression problem is by using correlation coefficients [32]. In this paper, we use *Spearman's correlation coefficient* (SCC) computed using the voltage phasors for feature selection for DNN-based DSSE. SCC can capture the correlation between nonlinear random variables whose behavior is monotonically increasing or decreasing, which is a common feature of voltages along a distribution feeder. It was observed that transformers/regulators and multiple outgoing laterals from the feeder head split the SCC matrix into multiple clusters. Hierarchical clustering is applied to the SCC matrix to find the group of nodes that can be monitored by one SMD. In hierarchical clustering the number of clusters is a hyperparameter that must be chosen in advance. We start with one cluster and add more clusters based on the budget constraint. The distance between the clusters (say, $r$ and $s$) is calculated using the Ward method [33], as shown below,

$$d(r,s) = \sqrt{\frac{2n_r n_s}{(n_r + n_s)}} \|\bar{x}_r - \bar{x}_s\|_2 \qquad (3)$$

where $\|\cdot\|_2$ is the Euclidean distance, $\bar{x}_r$ and $\bar{x}_s$ are the centroids of clusters $r$ and $s$, and $n_r$ and $n_s$ are the number of elements in clusters $r$ and $s$. Note that one SMD is placed in every cluster since adding more SMDs to the same cluster may not significantly reduce the estimation error of the overall system as the features in the same cluster are more correlated.

An overview of the integrated measurement selection algorithm is provided in Algorithm I. In this algorithm, Budget refers to the budget allocated for SMD placement, TI$_{accuracy}$ and DSSE$_{accuracy}$ are the minimum desired accuracy for TI and DSSE, respectively, DSSE$_{corr}$ is the minimum SCC between each pair of nodes, and $M$ is the number of nodes in the system.

## V. DATA CONSIDERATION & ALGORITHM IMPLEMENTATION

This section describes the steps that were followed to create the database required to train the two DNNs, while accounting for the unique characteristics of the distribution system and the attributes of the sensing system. To avoid repetition, the explanation is provided w.r.t. the DNN created for performing DSSE. Finally, the implementation procedure is described.

### A. Database Creation

As mentioned in Section I, smart meter measurements become available after a delay of at least a few hours, implying that they cannot be directly used for real-time DSSE. Therefore, the proposed methodology uses the historical slow timescale smart meter readings in the offline training process of the DNNs. The smart meter energy readings are first converted to average power by dividing the energy with the corresponding time interval. Then, the aggregated net injection at the distribution transformer level is calculated by summing up the readings of the smart meters connected to the transformer. The net load at each transformer is treated as a random variable.

| Algorithm I: Integrated SMD placement for DNN-based TI and DSSE |
|---|
| **Inputs:** Budget, TI$_{accuracy}$, DSSE$_{accuracy}$, DSSE$_{corr}$, $M$ |
| **Output:** Location of the SMDs |
| **A.  SMD placement for TI:** |
| A.i.   $N_{feature} = 1$ |
| A.ii.  If there are no switches in the system, go to (B) |
| A.iii. Apply sequential forward selection with $N_{feature}$ features to place SMDs |
| A.iv.  If SMD cost ≥ Budget, then End, else set $N_{feature} = N_{feature} + 1$ |
| A.v.   If TI$_{accuracy}$ is satisfied, then go to (B), else go to (A.iii) |
| **B.  SMD placement for DSSE:** |
| B.i.   $N_{cluster} = 1$ |
| B.ii.  Calculate SCC between each voltage phasor $V_{ij}^{kl}$ $\forall i \in \{A, B, C\}, \forall j \in \{\text{mag, ang}\}, \& \forall k, l \in \{1, ..., M\}$ |
| B.iii. If SCC $\forall k, l \in \{1, ..., M\}$ is greater than DSSE$_{corr}$ $\forall i \in \{A, B, C\}$ & $\forall j \in \{\text{mag, ang}\}$ then go to (B.vii.) |
| B.iv.  $N_{cluster} = N_{cluster} + 1$ |
| B.v.   Apply hierarchical clustering to each SCC matrix for $\forall i \in \{A, B, C\}$ & $\forall j \in \{\text{mag, ang}\}$ |
| B.vi.  Find common node in each cluster for each SCC and place SMD on this node |
| B.vii. If DSSE$_{accuracy}$ is satisfied or SMD cost ≥ Budget, then End, else go to (B.iv.) |

Next, a *kernel density estimator* (KDE) is used to learn the (non-Gaussian) distribution of aggregated smart meter readings. Although KDE is suitable for learning the PDF of data samples that do not follow a parametric PDF, it is prone to overfitting which causes loss of generality of the fitted PDF [34]. To overcome this problem, we modify the KDE by adjusting its bandwidth to achieve 95% confidence interval ensuring that the fitted PDF effectively represents net load behavior. After the PDF of active power injection is obtained, the reactive power is computed by selecting a power factor from a uniform distribution lying between 0.95 and 1. Monte Carlo (MC) sampling is done next to pick active and reactive power injections from the learnt distribution to run a large number of power flows. The voltage and current phasors obtained from the solved power flows are used to create the training database.

### B. Time Resolution Difference between Smart Meter & SMD

During the online operation, the trained DNN is fed with streaming data from SMDs to perform DSSE at SMD timescales (sub-second time interval). However, as SMD data was not used in the training process, a question arises regarding the effectiveness of using calculated average power to represent instantaneous power injections at sub-second time resolution. In fact, it is statistically not possible to obtain the instantaneous power injections from the average power for one particular time interval. However, we hypothesize that if sufficient historical data for average powers is available, it is possible to approximate the PDF of instantaneous power injections using the historical average power measurements.

We perform two statistical tests, namely, the two sample Kolmogorov-Smirnov (KS) test and the Mann-Whitney (MW)

U-test, two confirm our hypothesis. The two sample KS test is a non-parametric test that examines the null hypothesis that the data in set 1 and set 2 are from the same distribution. The MW U-test is a non-parametric test that examines the null hypothesis that the data in set 1 and set 2 are from the distribution with the same median. Using these two statistical tests, the PDFs of instantaneous power injections and average power consumption are compared in terms of *shape* and *median*. If the null hypothesis is not rejected for both tests, it indicates that the PDF created based on historical average power consumption data can reliably approximate the PDF of instantaneous power injections. It should be noted that for performing these tests instantaneous power injections should be available for a particular time and distribution system. Subsequently, inferences drawn from these tests can be extended to other distribution systems in which the instantaneous power injections are not available (see Table II and its explanation).

### C. Embedding Unique Characteristics of Distribution System and Sensing System Attributes into DNN Training

The salient characteristics of the distribution system that were included in the physical network model used for creating the samples for DNN training are: wye-delta loads, zero-injection phases, distributed loads, single, double, and three-phase laterals, voltage regulators, transformers, and capacitor banks. Moreover, unlike transmission systems that are usually balanced, distribution networks are unbalanced; hence, DSSE was carried out for each phase separately. We modified the DNN architecture shown in Figure 1 to create separate neurons and layers for each phase as shown in Figure 2. However, measurements of all phases were fed into each block to account for the mutual coupling that exists between the phases.

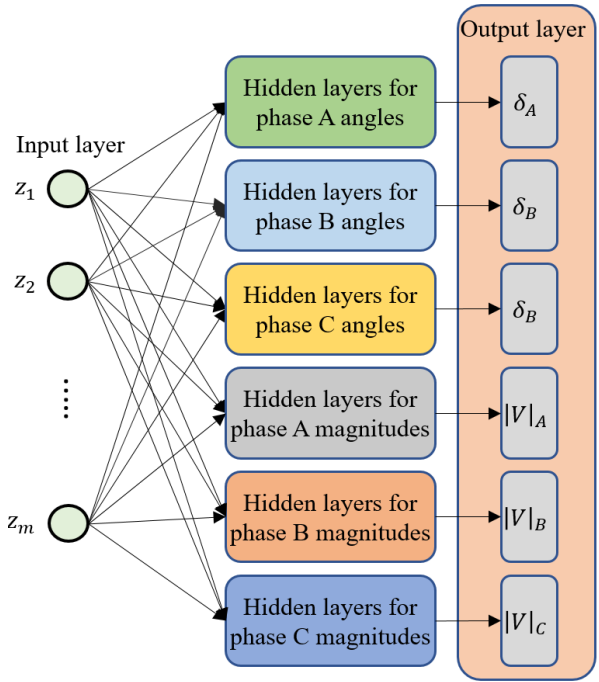

Figure 2: Modified DNN architecture for unbalanced distribution networks

An SMD has six channels which measures three nodal voltage phasors and three branch current phasors [3], providing real-time observability of the individual phases of the node where it is placed. To account for this attribute of SMDs during DNN training, each phasor magnitude and angle is treated as separate features that are fed into the input layer of the DNN. The voltage magnitude and angle of every phase of every node is estimated at the output layer (see Figure 2).

In order to replicate actual SMD measurements, appropriate measurement error must be added to the error-free voltages and currents obtained from the power flow solution (see Section V.A). According to the IEEE Standard [35], SMDs should meet the total vector error (TVE) requirement of 1%. It has also been widely assumed that the errors in SMD data follow a Gaussian distribution. However, SMDs are connected to the grid through an instrumentation channel consisting of instrument transformers, cables, and burden. These components not only cause the total measurement error to go beyond the 1% TVE limit, but also change the shape of the error distribution, e.g., from a Gaussian to a 3-component Gaussian mixture model (GMM) [6]. Note that the instrument transformer alone for voltage magnitudes, voltage angles, current magnitudes, and current angles can be as high as $\pm 1.2\%$, $\pm 1°$, $\pm 2.4\%$ and $\pm 2°$, respectively [36]. To account for these practical constraints, we proposed a two-level error model in [37], which is also used here. In the first level, the instrumentation channel error is modeled by a 3-component GMM with the corresponding magnitude and angle errors added to the error-free voltages and currents. In the next level, a Gaussian TVE is added to the previously obtained erroneous measurements. This two-level error model ensures generation of realistic SMD data.

### D. Implementation of DNN-based TI and DSSE

The procedure to be followed for implementing the proposed methodology is presented in Figure 3. The model is split into an offline learning stage and a real-time operation stage.

#### 1) Offline learning

In the offline learning stage, it is assumed that the distribution system is equipped with smart meters at all the nodes and the data produced from these meters have been saved for some period of time (e.g., 1-year). This historical smart meter data is used to find the PDFs of the power injections at a given node. MC sampling of the active and reactive power injections is done from the best-fit PDF to run three-phase unbalanced power flows (the power flow solution process is explained in the Appendix). After a power flow is solved, for a given value of active and reactive power injections at all the nodes, the voltage and current phasors of the nodes equipped with SMDs (with added measurement noises) are used as the input of the DNN for DSSE, while the voltage phasors of all the nodes are used as the output of the DNN for DSSE. In addition, topology information is saved for each solved power flow to train the DNN for TI. For the current configuration of the system (called base topology), a DNN is trained for performing DSSE. For other network configurations, all voltage phasors and SMD data are saved for each feasible topology. Once the data from power flow results are saved for all feasible topologies, a separate DNN is trained for performing TI.

#### 2) Real-time operation

For real-time operation, the trained DNN-based TI is used to estimate the current network topology from real-time SMD data. If the estimated topology is consistent with the base topology, the DNN trained for the base topology is employed to perform DSSE. If the estimated topology is different from the base topology, Transfer learning (via fine-tuning) is used to update the DNN used for performing DSSE, and the current topology becomes the new base topology. In summary, the DNN trained for TI does not need to be updated for different topologies as it is trained for all (feasible) topologies and can therefore estimate the current topology in real-time. When the network topology does change, only the DNN trained for DSSE must be updated in real-time using fine-tuning.

## VI. SIMULATION RESULTS

### A. Simulation Settings

#### 1) Distribution system setup

Simulations are performed on a radial IEEE 34-node system (System S1) [38] and a meshed 240-node distribution network of Midwest U.S. (System S2) [39]-[40]. In System S1, three distributed generation (DG) units having ratings of 135kW, 60kW, and 60kW are also placed on nodes 822, 848, and 860, respectively, to model the effect of renewable generation. The loads and DG units are varied based on the Pecan Street historical data [41] to create different scenarios for this system. System S2 has smart meters installed at customer premises and all the characteristics of a modern distribution network, such as underground and overhead lines, capacitors and voltage regulators, and single, double, and three-phase laterals and loads. One-year of smart meter readings is also available for this system. PDFs (computed using KDE) were fit to the historical hourly smart meter data. Network models of Systems S1 and S2 are available in OpenDSS [42]. The budget constraint was set at two SMDs for System S1 and ten SMDs for System S2. The $TI_{accuracy}$ was set at 95%, while the $DSSE_{accuracy}$ was set at 0.15° for phase angle mean absolute error (MAE) and 0.30% for voltage magnitude mean absolute percentage error (MAPE). While performance of DNN-based DSSE is evaluated for both systems, DNN-based TI results are presented for System S2 only, as System S1 does not have any switches.

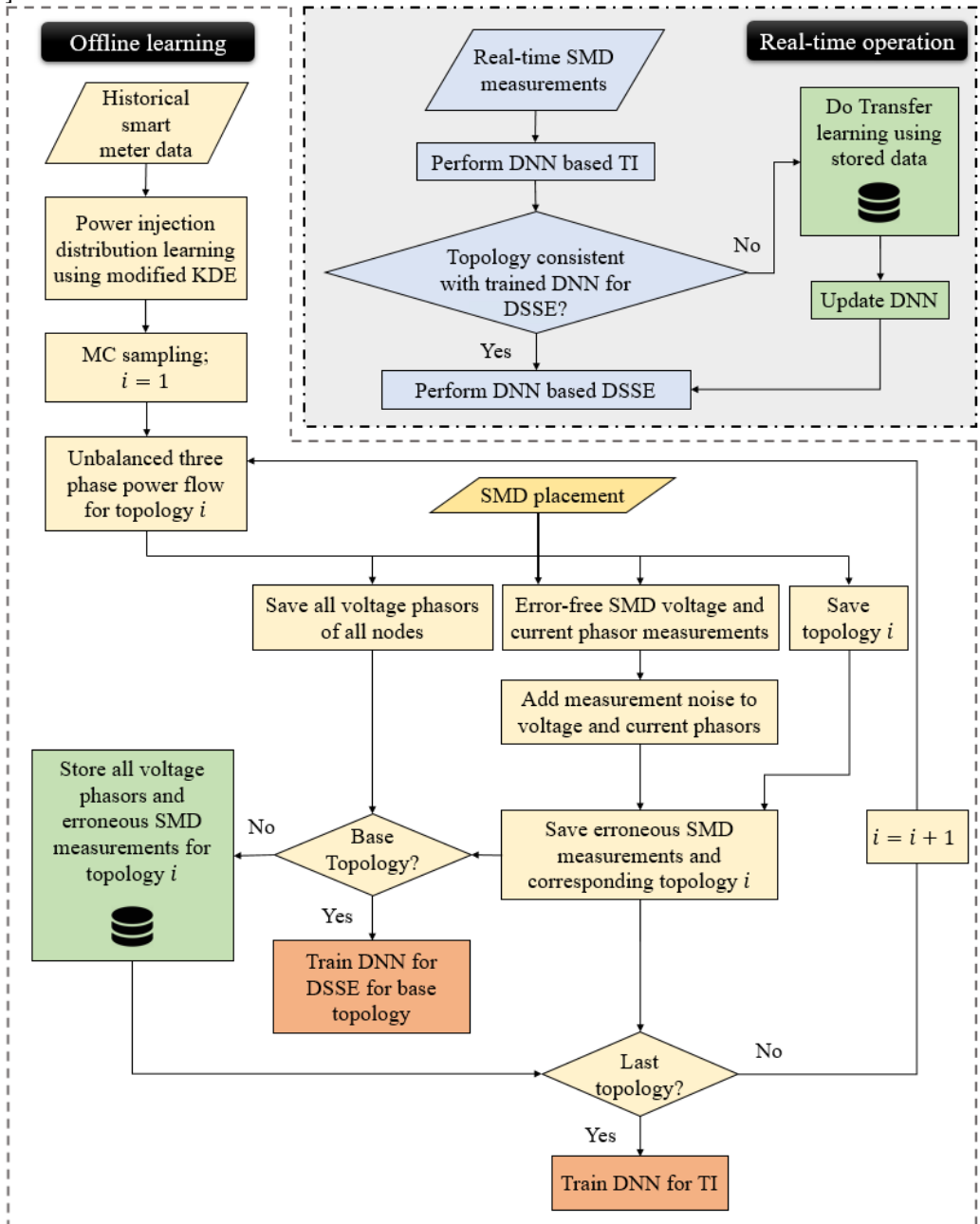

Figure 3: Implementation of the proposed DNN-based TI and DSSE

### 2) Neural network setup

The hyperparameter information for the three DNNs found using the Random search method explained in Section II.C is summarized in Table I. The Random search was implemented using the WANDB toolbox [43]. The search space for each hyperparameter was identified based on our familiarity with DNNs and existing literature on DNN for power system applications (e.g., [13]). It should be noted that the number of neurons in each hidden layer and the number of hidden layers for DNN-based DSSE correspond to each block shown in Figure 2. TensorFlow v.2.3.0 was used in Python v.3.8 to carry out the training. All simulations were performed on a computer with 256.0 GB RAM, Intel Xeon 6246R CPU @3.40GHz, Nvidia Quadro RTX 5000 16 GB GPU.

**Table I: Hyper-parameters for DNN-based TI and DSSE**

| Hyper-parameters | DNN-based TI for S2 | DNN-based DSSE | |
|---|---|---|---|
| | | **S1** | **S2** |
| **No. of neurons in input layer** | 2×No. of measured phasors by all SMDs | 2×No. of measured phasors by all SMDs | |
| **No. of neurons in each hidden layer** | 800 | 200 | 500 |
| **No. of hidden layers** | 5 | 5 | |
| **No. of output neurons** | No. of feasible topologies | No. of states | |
| **Hidden layer activation function** | ReLU | ReLU | |
| **Output layer activation function** | SoftMax | Linear | |
| **Initializer method** | He normal | He normal | |
| **Optimizer** | ADAM | ADAM | |
| **No. of epochs** | 50 | 1000 | |
| **No. of samples** | 1,000 per topology | 12,500 | |
| **Training percentages** | 80% training and validation, 20% testing | 80% training and validation, 20% testing | |
| **Learning rate (lr)** | 0.02726 with reduce learning rate on Plateau | 0.09456 | 0.0988 |
| | | with reduce learning rate on Plateau | |
| **Regularization** | 30% Dropout | 50% Dropout | 50% Dropout |
| **Loss function** | Categorical cross-entropy | Mean squared error | |

### B. IEEE 34-Node System (System S1)

As historical smart meter data was not available for System S1, Pecan Street data was used to generate realistic loading scenarios for this system. Secondly, quarterly, and hourly smart meter data is available for 25 houses in the Pecan Street dataset. To compute aggregate loading at the distribution transformer level, the power consumption of 6-8 randomly chosen houses, were added. Then, KS and MW U-tests were performed between instantaneous secondly power and hourly average power (Scenario 1), and between instantaneous secondly power and quarterly average power (Scenario 2), for both loads and DGs. The number of times that the null hypothesis was rejected for each test are shown in Table II. It can be realized from the table that the medians of the PDFs of average and instantaneous power were the same as the MW U-test was not rejected in any scenario. Similarly, the KS test results show that the shape of the two PDFs for instantaneous and average power measurements was the same in more than 96% of the scenarios. This implies that PDFs created from quarterly or hourly historical smart meter data can be used to reasonably approximate the PDF of instantaneous power injections.

**Table II: Statistical test results for Pecan Street data**

| Data type | | KS test [%] * | MW test [%] * |
|---|---|---|---|
| **Load** | Scenario 1 | 0.8 | 0 |
| | Scenario 2 | 0.15 | 0 |
| **DG**** | Scenario 1 | 3.9 | 0 |
| | Scenario 2 | 0 | 0 |

\* Percentages are calculated based on 1,000 MC samples
\*\* DG refers to aggregated rooftop solar PV generation at the distribution transformer level for varying weather conditions across multiple days

Figure 4 shows the DNN-based DSSE performance for voltage angle and magnitude estimation of phase A in terms of MAE and MAPE. For hourly and quarterly errors, historical data was used for both training and testing (blue bars). Red bars show the DNN performance when training was done based on the PDF of hourly and quarterly data and testing was done based on the scenarios where secondly instantaneous injections were used. The difference in the heights of the blue and red colored bars are an indication of the unavoidable errors that will be present if the PDFs are generated from slow timescale (hourly and quarterly) measurements instead of instantaneous (secondly) power injections. Thus, Figure 4 is the quantification of the inferences drawn from Table II in the context of DNN-based DSSE.

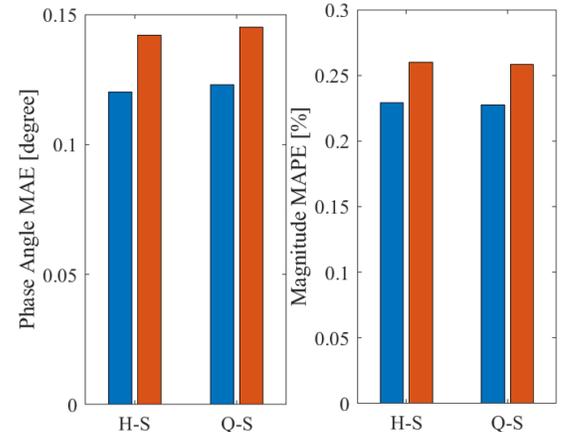

Figure 4: Comparison of DNN-based DSSE performance between Hourly (H), Quarterly (Q) and Secondly (S) data

There are no switches present in System S1, so only measurement selection for DNN-based DSSE is required. First SCCs for voltage phasors were calculated. Subsequently, hierarchical clustering was applied six times to SCC matrices for each phase (A, B, C) and each type of state variable (magnitude and angle). The cluster number that each node belongs to is shown in Figure 5. It can be observed from the figure that regulator R2 splits the system into two clusters. For instance, dark blue square shows that all the nodes before R2 are grouped in cluster 1 in terms of voltage magnitudes of phase A. Similarly, light blue squares indicate that all the nodes after R2 belong to cluster 2 for voltage magnitudes of phase A. This implies that the *minimum number of SMDs required to perform DNN-based DSSE is two* (one in each cluster). Note that all nodes in each cluster are potential candidates for placing a SMD. However, the starting node in each cluster is the most effective location as it captures the total current entering the cluster [11]. Therefore, nodes 800 and 832 were chosen for SMD placement for cluster 1 and cluster 2, respectively.

Figure 6 shows the MAE and MAPE of phase C for DNN-based DSSE for four cases. Case (a): one SMD is placed inside cluster 1 (800-802, red dots). Case (b): one SMD is placed in cluster 2 (832-858, blue dots). Case (c): two SMDs are placed at two locations (800-802and 828-830) that belong to the same cluster (yellow dots). Case (d): two SMDs are placed at two locations (800-802 and 832-858) that belong to two different clusters (green dots). Note that a location *i-j* means that the SMD monitors the voltage at node *i* and the currents flowing from node *i* to node *j*. From Figure 6, it is clear that the overall error decreased when the two SMDs were placed in two different clusters (Case (d)), which is consistent with the logic proposed in Section IV.B. Similar results were also obtained for the angles and magnitudes of the other phases. Some additional remarks regarding this figure are provided below.

*Remark* 1: The phase angle MAE profile is relatively flat. This is because the SCC values for the phase angles of all the nodes are close to 1, implying that the intercorrelation between the clusters is very high. As such, placing one SMD in either cluster is able to lower the angle MAE of the nodes belonging to both the clusters. Furthermore, adding more SMDs is able to lower the angle MAE of all the nodes as more information/features are provided to the DNN (for DSSE).

*Remark* 2: The voltage magnitude MAPE profile shows the impact of SMD placement more clearly. There is a significant difference in the SCC values between nodes belonging to the two clusters. This is caused by regulator R2, which greatly lowers the correlation between the clusters. As a result, placing one SMD in one cluster has little impact on the MAPE of the other cluster. A similar behavior would have been observed for the angle MAE profile as well, if R2 was equipped with a phase shifter; however, that is not the case for System S1. Lastly, as Algorithm I accounts for both phase angles and magnitudes, it will place sensors that will lower estimation errors across the entire system while remaining within the budget constraint.

*Remark* 3: Regulator R1 and the transformer between nodes 832 and 888 also impacted the DSSE performance (particularly, the magnitude MAPE profile) of System S1. However, more SMDs could not be added to this system as the budget constraint had been hit. Furthermore, for obtaining the results shown in Figure 6, the real-time knowledge of the tap settings of the regulators and transformers was not required. This is because during the offline learning stage, the tap settings were automatically adjusted [44] based on loading and feeder head voltage scenarios. Hence, the DNN (for DSSE) became aware of the effects of different tap settings during the training itself.

*Remark* 4: By comparing Figure 4 and Figure 6, we can identify two factors that can increase the estimation error: (1) Using historical smart meter energy readings to approximate PDFs of instantaneous power injections. (2) SMD placement. The error caused by the former is unavoidable as it is statistically not possible to create the exact instantaneous power injections from the average value provided by smart meters. However, the error caused by the latter is controllable, and can be reduced by intelligently placing SMDs.

*Remark* 5: The number of measurements provided by the two SMDs installed in System S1 is 24 (3 phasor measurements of voltage and current, respectively, by each SMD), while the number of states (voltage magnitude and angle) to be estimated, considering all phases in single and three phase nodes, is 172. As 172>>24, it can be realized that the proposed methodology is able to perform time-synchronized DSSE for System S1 when it is incompletely observed by SMDs in real-time.

Next, the performance of DNN-based DSSE is compared with linear state estimation (LSE) [45]; the results are shown in Table III. To satisfy the complete real-time observability requirement of LSE, System S1 needed 26 SMDs (based on the optimization framework proposed in [46]). It can be observed from Table III that DNN-based DSSE gives similar results as classical LSE in terms of both angle MAE and magnitude MAPE with only *two* SMDs, validating the outcome of Algorithm I for this system. Note that the LSE results correspond to a purely Gaussian noise of 1% TVE and no measurement redundancy. The results of LSE (and DNN-based DSSE) can be further improved if a smaller TVE is considered.

Furthermore, to give additional confidence in the results, the *tolerance interval* was computed. The tolerance interval provides the upper and/or lower bounds within which, with some confidence level, a specified proportion of the samples fall [47]. The confidence level and population proportion were both set at 95%; i.e., the upper bound of the tolerance interval was used as a measure of the confidence of the state estimates. In reference to Table III, 0.4° tolerance interval for voltage angles implies that with a confidence level of 95%, 95% of the error values in estimating the angles by the DNN-based DSSE were less than 0.4°. It can also be observed from Table III that the proposed DNN-based DSSE was robust to non-Gaussian noise as its performance deteriorated negligibly when the two-level error model was used.

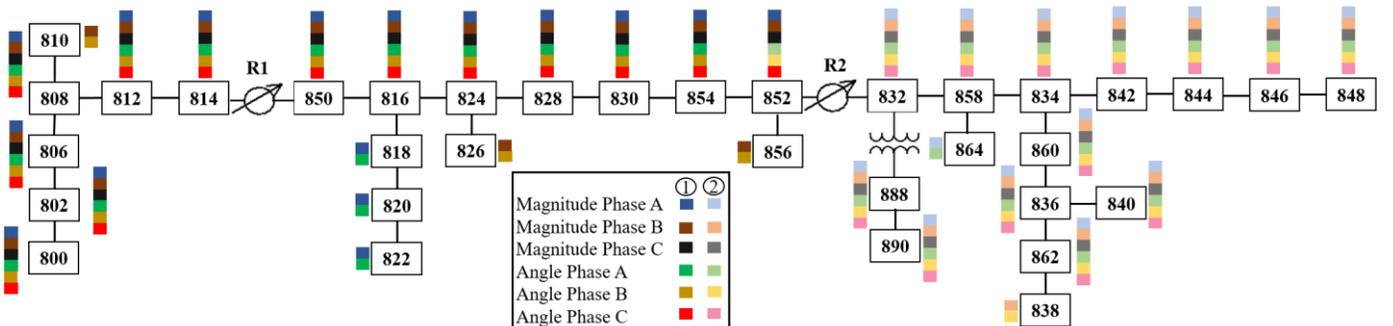

Figure 5: Clustering results for SMD placement for DNN-based DSSE for System S1

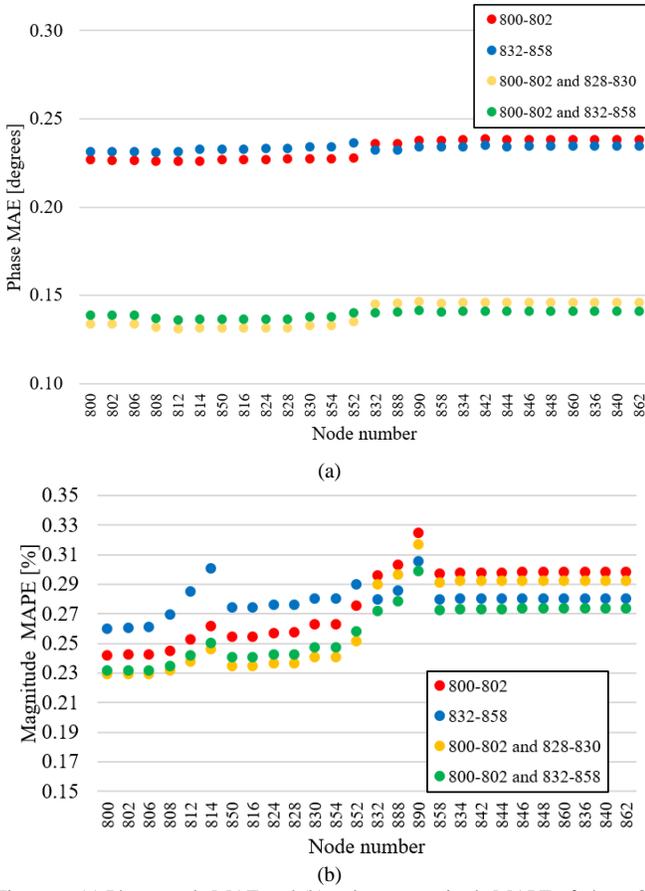

(a)

(b)

Figure 6: (a) Phase angle MAE and (b) voltage magnitude MAPE of phase C for DNN-based DSSE for System S1

In addition to the comparison between DNN-based DSSE and LSE, support vector regression (SVR) with a polynomial kernel was also used to perform DSSE for System S1 to provide a comparative study between two ML-based techniques. We can see from Table III that DNN-based DSSE outperforms SVR-based state estimation, highlighting the superior performance of the proposed methodology.

Lastly, to demonstrate the ability of the proposed approach to provide real-time state estimates, a stream of high-speed data obtained from the Pecan Street dataset were set as inputs to the trained DNN for DSSE. The DNN was able to consistently track the variations in the states (see plot of Phase C voltage angle in Figure 7). Moreover, the DNN took only 0.01 seconds to produce the estimates. This is because a trained DNN performs a matrix multiplication of the input values with the weights and biases of its neurons – a process that can be done very fast. Thus, this study shows that the proposed approach can provide fast (sub-second) situational awareness to distribution systems that are incompletely observed by SMDs in real-time.

### C. 240-Node Network of Midwest U.S. (System S2)

#### 1) DNN-based TI and DSSE

Due to switches being present in System S2, SMD placement for TI was done first based on the integrated measurement selection algorithm (see Section IV). Considering the locations of the 9 switches (see Figure 8), 84 feasible topologies were identified. 1,000 samples were generated by varying the loads for each of the 84 topologies. Based on the sequential forward selection algorithm, 4 SMDs were placed at 1010-2057, 2012-2013, 2021-2026, and 3030-3031 to attain a TI accuracy of 99.19%. The locations are depicted by green ovals in Figure 8.

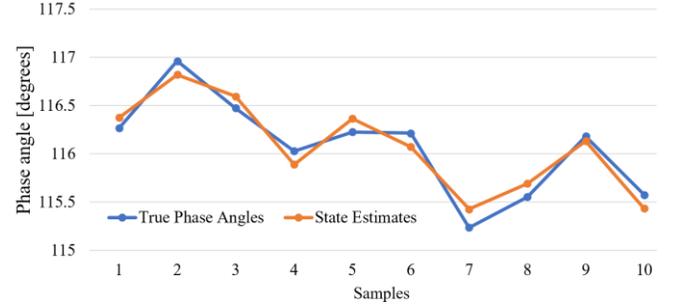

Figure 7: State estimates for Phase C voltage angle along with true values for node 846 of System S1

**Table III: Comparing the performance of DNN-based DSSE with classical LSE for System S1**

| Method | Error model | Phase error [degrees] | | Magnitude error [%] | | # SMD |
|---|---|---|---|---|---|---|
| | | MAE | Tolerance interval | MAPE | Tolerance interval | |
| LSE | 1% Gaussian TVE | 0.14 | 0.40 | 0.25 | 0.60 | 26 |
| DNN-based DSSE | 1% Gaussian TVE | 0.13 | 0.40 | 0.26 | 0.61 | 2 |
| | Two-level GMM | 0.14 | 0.42 | 0.26 | 0.62 | 2 |
| SVR | Two-level GMM | 0.19 | 0.50 | 0.31 | 0.72 | 2 |

Measurement selection for DNN-based DSSE using the SCC was investigated next. It was observed that based on this metric, System S2 could be split into five clusters: one comprising Feeders A, two comprising feeder B, and two comprising Feeder C; implying that at least five SMDs would be required. However, four SMDs (= one in Feeder A, two in Feeder B, and one in Feeder C) had already been placed in this system based on the measurement selection algorithm for DNN for TI. These four SMDs satisfied the requirements for DNN-based DSSE for three clusters. Thus, two more SMDs at 2044-2053 and 3118-3107 were added to complete the SMD placement for DNN-based DSSE for this system.

The performance of DNN-based DSSE was now compared with LSE and SVR-based state estimation for System S2. The total number of SMDs required for complete real-time observability of System S2 was 113 (based on the optimization framework proposed in [46]). It can be observed from Table IV that the DNN-based DSSE gives similar results as LSE and outperforms SVR-based state estimation with only six SMDs. Moreover, the accuracy of DNN-based DSSE is practically the same with 1% Gaussian TVE and with the two level-GMM error model, confirming that the DNN-based DSSE is robust against both the noise model and the noise magnitude.

Lastly, note that the number of states to be estimated for System S2 is 924, while the number of measurements obtained from the six SMDs is 72 (<<924). These observations, along with the analyses conducted in the previous sub-section, confirm that the proposed approach can successfully perform time-synchronized DSSE for distribution systems that are incompletely observed by SMDs in real-time. It should be noted that the DSSE results obtained in Table IV and the achieved TI accuracy of 99.19% are based on the integrated SMD placement strategy presented in Algorithm I. If SMD placement targets only one task (DSSE or TI), the performance of the other task will deteriorate. This is realized from Table V which compares the DSSE and TI performance for integrated and non-integrated SMD placement. It can be seen from Table V that for System S2, the minimum number of SMDs required to only achieve requisite $TI_{accuracy}$ is four. However, the MAE for DSSE becomes $0.17°$ with four SMDs, which is higher than the pre-specified $DSSE_{accuracy}$ threshold of $0.15°$ (see Section VI.A.1). This increase in error from $0.15°$ to $0.17°$ primarily occurred in the second cluster of Feeder C, in which during the measurement selection for DNN for TI, no SMD was placed. Similarly, the minimum number of SMDs required to only achieve requisite $DSSE_{accuracy}$ is five. However, this decreases the TI accuracy to 80.27%, which is less than the threshold set for $TI_{accuracy}$. The integrated approach presented in Algorithm I picks *six nodes as the minimum number of locations* where SMDs must be placed for System S2. The last row of Table V confirms that this solution is able to satisfy the requisite $TI_{accuracy}$ and $DSSE_{accuracy}$, simultaneously. Finally, note that adding more than six SMDs will have minimal effect on both TI and DSSE accuracy (due to the reasons already mentioned in *Remark 2* and explanation of Figure 6 for System S1), especially considering the hard budget constraints typically associated with placing SMDs in distribution systems.

**Table IV: Comparing the performance of DNN-based DSSE with classical LSE for System S2**

| Method | Error model | Phase error [degrees] | | Magnitude error [%] | | # SMD |
|---|---|---|---|---|---|---|
| | | MAE | Tolerance interval | MAPE | Tolerance interval | |
| LSE | 1% Gaussian TVE | 0.14 | 0.35 | 0.25 | 0.61 | 113 |
| DNN-based DSSE | 1% Gaussian TVE | 0.15 | 0.36 | 0.25 | 0.60 | 6 |
| | Two-level GMM | 0.15 | 0.37 | 0.26 | 0.62 | 6 |
| SVR | Two-level GMM | 0.18 | 0.40 | 0.31 | 0.71 | 6 |

**Table V: Comparison of DSSE and TI performance for integrated and non-integrated SMD placement for System S2**

| Placement target | SMD locations | DSSE Performance | | TI accuracy [%] |
|---|---|---|---|---|
| | | Angle MAE [degrees] | Magnitude MAPE [%] | |
| SMDs for TI only | 1010-2057; 2012-2013; 2021-2026; 3030-3031 | 0.17 | 0.31 | 99.19 |
| SMDs for DSSE only | 1010-2057; 2012-2013; 2044-2053; 3030-3031; 3107-3118 | 0.15 | 0.26 | 80.27 |
| Integrated | 1010-2057; 2012-2013; 2021-2026; 2044-2053; 3030-3031; 3118-3107 | 0.15 | 0.26 | 99.19 |

### 2) Transfer learning for different network topologies

When topology changes occur, after correctly identifying the new topology using DNN-based TI, the DNN trained for doing DSSE for the old topology, must be updated. As mentioned in Figure 3, the TI and DSSE work sequentially, and Transfer learning is used to update the DNN for DSSE after the topology of the system changes. Four different topologies are considered below to show the ability of the proposed approach in handling different system configurations. Initially, the system is operating in the base topology, T1, which is radial. Next, status of three switches are changed to create a meshed network, described by T2. Then, configurations of five switches are changed to create a new topology, T3. Finally, in the fourth step, T3 changes to another topology, T4, which is different from all the previous topologies. The summary of network reconfigurations is shown in Table VI.

Figure 9 presents the results for topology changes and its impact on DSSE with and without Transfer learning. It can be seen from the plots that it takes *about 1 minute* for the fine-tuning of the DNN, while complete training for a new topology would have taken *two hours*. This is because 10,000 samples and 1,000 epochs were needed for training and validation of a completely new DNN for DSSE for a new topology (see Table I), while by taking advantage of fine-tuning only 3,000 samples and 32 epochs were needed; thereby reducing the training time drastically. This is an important result because if different switching events were to manifest every few minutes, then without Transfer learning we will not be able to achieve fast and accurate DSSE results when it is needed the most. Hence, this quick update of the DNN-based DSSE considerably improves the real-time monitoring capability of the proposed approach during switching events.

**Table VI: Switch configurations for different topologies**

| Switch name | Network reconfiguration | | | |
|---|---|---|---|---|
| | T1 ⟹ T2 | | T3 ⟹ T4 | |
| CB_101 | 1 | 1 | 0 | 0 |
| CB_102 | 0 | 1 | 1 | 1 |
| CB_201 | 1 | 1 | 0 | 1 |
| CB_202 | 1 | 0 | 1 | 1 |
| CB_203 | 1 | 1 | 1 | 1 |
| CB_204 | 0 | 1 | 1 | 0 |
| CB_301 | 1 | 1 | 1 | 1 |
| CB_302 | 1 | 1 | 0 | 1 |
| CB_303 | 0 | 0 | 1 | 1 |

Lastly, the angle MAE results are now compared with and without fine-tuning of the DNN for DSSE. It is observed from Figure 9 that if the old DNN for DSSE (created for T1) was used for the new topologies (T2, T3, T4), the error can increase by more than 1.5 times for the change from T1 to T2, more than two times for the change from T1 to T3, and more than three times for the change from T1 to T4 (compare heights of the orange bars and blue bars in Figure 9), respectively. However, the state estimator performance is similar for fine-tuning and complete training (compare heights of green bars and blue bars in Figure 9). Therefore, by using Transfer learning, DNN-based DSSE can be done quickly and accurately during varying network topologies.

## VII. CONCLUSION

In this paper, a DNN framework for performing *unbalanced three-phase time-synchronized DSSE for different network configurations* is proposed that does not require complete network observability by SMDs in real-time. The unique feature of the proposed algorithm is that it neither relies on forecasted/pseudo-measurements nor does it use slow timescale AMI data directly for DNN training. Instead, historical data is used to find a mapping between the states (voltage phasors) and the SMD measurements, with the mapping being realized using a DNN. When a change in topology occurs, the proposed framework first detects the change using a DNN built for topology identification, and subsequently employs Transfer learning via fine-tuning to update the DNN for DSSE in real-time for the new topology.

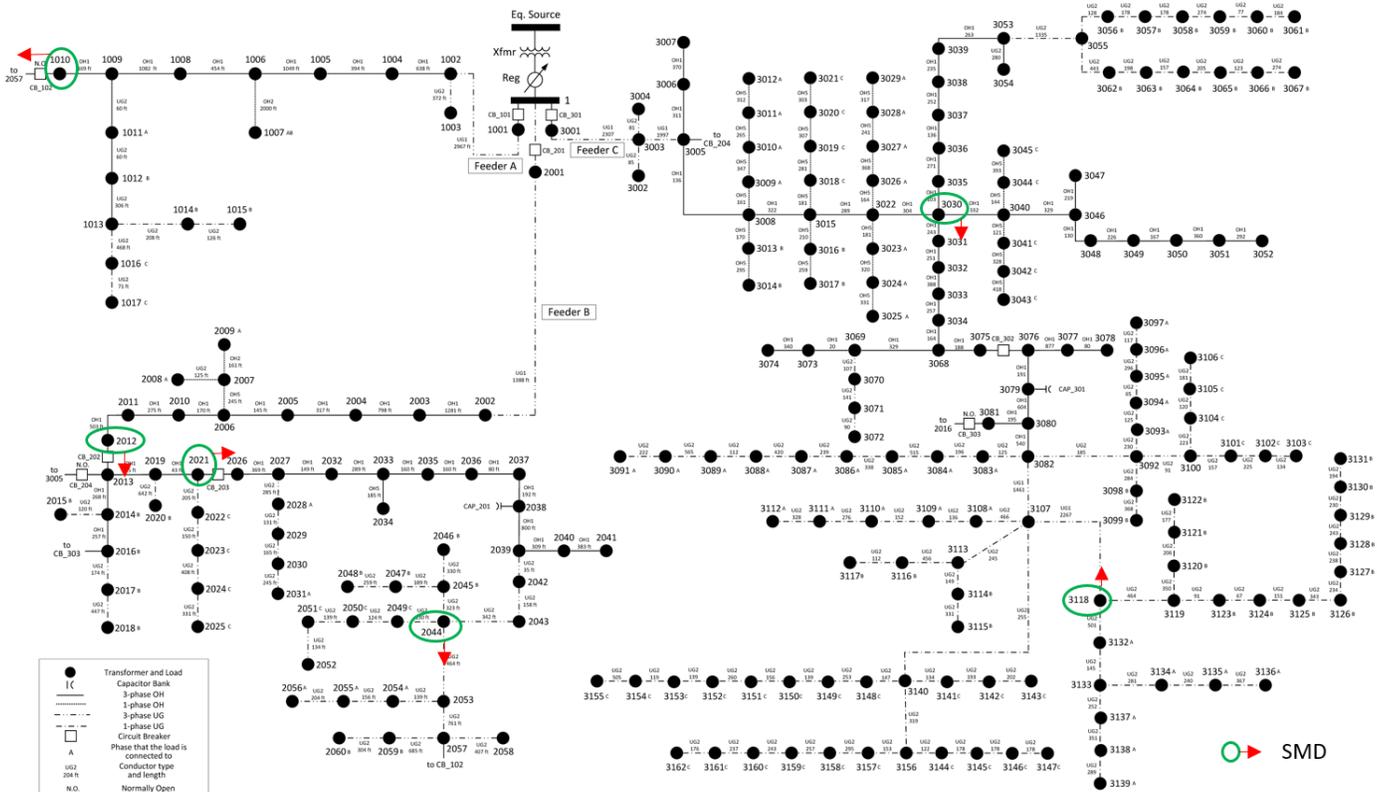

Figure 8: System S2 with SMD locations

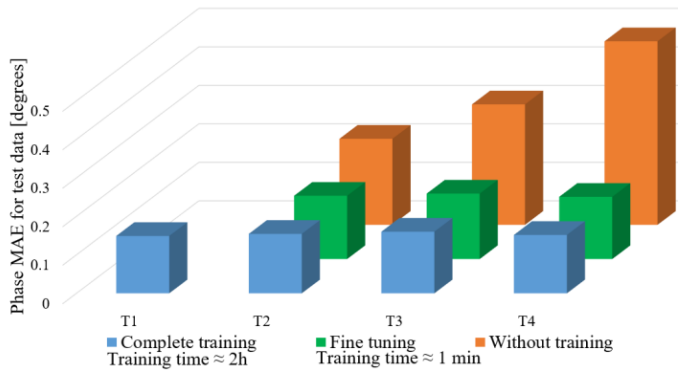

Figure 9: Comparative study of DNN-based DSSE with and without fine-tuning of the DNN

A detailed methodology for SMD placement is also proposed to enhance the performance of DNN-based DSSE for varying network configurations. Being a greedy search method, this measurement selection strategy for SMD placement is not guaranteed to be optimal. However, it is deemed acceptable for the following reasons: (i) the problem being solved here is unsolvable in the classical sense (i.e., it has larger number of unknowns than knowns), therefore, there may not be a sensor placement algorithm that consistently gives the best results under all operating conditions, and (ii) the focus is not on *optimizing* sensor placement but on getting reasonable topology identification and state estimation results, which the proposed placement strategy is able to provide.

The performance of the proposed DNN-based DSSE is validated by comparing it with the classical LSE as well as another ML-based state estimator (SVR-based state estimation). The simulation results on a renewable-rich IEEE 34-node distribution feeder and the meshed 240-node Midwest U.S. system show that the proposed method: (1) can achieve similar DSSE accuracy with a significantly smaller number of SMDs, (2) can efficiently detect varying network topologies for reconfigurable distribution systems, (3) ensures reliable DSSE for different topologies, and (4) is robust against non-Gaussian measurement noise, non-parametric load variations, and renewable energy fluctuations. The ability of the proposed algorithm to provide reliable state estimates with very few SMDs in large distribution networks for different topologies makes it a suitable candidate for enhanced monitoring, protection, and control of actual distribution systems.

## APPENDIX

In the scenario generation process, OpenDSS [42], which is a distribution analysis software, was used to solve power flow cases for different scenarios; here scenario refers to different values of active and reactive power injections at all nodes of the network. OpenDSS uses Forward-Backward sweep algorithm to calculate the voltage phasors of the system. Forward-Backward sweep is an iterative algorithm based on Kirchhoff's circuit laws [48]. Forward-Backward sweep method comprises three steps for a distribution network shown in Figure 10.

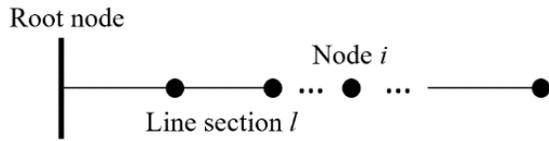

**Root node**

**Node $i$**

**Line section $l$**

Figure 10: Distribution system

The voltage phasor at root node is assumed to be known and the initial voltage for all the other nodes are assumed to be equal to the root node voltage.

In Step 1, three phase nodal currents are calculated as follows:

$$I_{abc,i} = \left( S_{abc,i} \odot \frac{1}{V_{abc,i}^{(k-1)}} \right)^* - Y_{abc,i}^* V_{abc,i}^{(k-1)} \tag{A.1}$$

where $I_{abc,i}$ is a vector of three phase current injections at node $i$, $S_{abc,i}$ is a vector of three phase power injections at node $i$ and $V_{abc,i}^{(k-1)}$ is a vector of three phase voltages at node $i$ at iteration $k$-1. $Y_{abc,i}$ is a diagonal matrix comprising admittance of all shunt elements at node $i$. $\odot$ denotes element-wise multiplication of vectors.

In Step 2 backward sweep is done to sum up line section currents starting from the last line along the feeder towards the root node. The current in line $l$ is:

$$J_{abc,l}^{(k)} = -I_{abc,j}^{(k)} + \sum_{m \in M} J_{abc,m}^{(k)} \tag{A.2}$$

where $J_{abc,l}$ are the current flows in line section $l$, and $M$ is the set of line sections connected to node $j$.

In Step 3 forward sweep is done to update voltages at all nodes. Starting from the root node and moving towards the end of the feeder. The voltage at node $j$ is:

$$V_{abc,j}^{(k)} = V_{abc,i}^{(k)} - Z_l J_{abc,l}^{(k)} \tag{A.3}$$

where $Z_l$ is a $3 \times 3$ matrix containing self and mutual impedances of three phases.

After these three steps are run in one iteration, the power mismatch at each node for all phases are calculated using (A.4).

$$\Delta S_{abc,i}^{(k)} = V_{abc,i}^{(k)} \odot \left( I_{abc,i}^{(k)} \right)^* - Y_{abc,i}^* |V_{abc,i}|^2 - S_{abc,i} \tag{A.4}$$

If the real or imaginary part (real or reactive power) of any of these power mismatches is greater than the convergence criterion, Steps 1, 2, and 3 are repeated until convergence is achieved.